\pdfoutput=1

\documentclass[11pt]{article}

\usepackage{graphicx}
\usepackage[table,dvipsnames]{xcolor}
\usepackage{pifont}
\newcommand{\cmark}{\ding{51}}
\newcommand{\xmark}{\ding{55}}

\usepackage{multirow}
\usepackage{amsmath,bm}
\usepackage{mathtools}
\usepackage{booktabs}
\usepackage{amsthm}
\usepackage{amssymb}
\usepackage[toc,page]{appendix}

\DeclareMathOperator*{\argmax}{arg\,max}

\newcommand\mycdots{\hbox to 0.75em{$\cdot$\hss$\cdot$\hss$\cdot$}}

\usepackage{acl}

\usepackage{times}
\usepackage{latexsym}
\usepackage[T1]{fontenc}
\usepackage[utf8]{inputenc}
\usepackage{microtype}
\usepackage{inconsolata}

\title{VOLTA: Improving Generative Diversity by Variational Mutual Information Maximizing Autoencoder}

\author{Yueen Ma\textsuperscript{\rm 1},
        Dafeng Chi\textsuperscript{\rm 2},
        Jingjing Li\textsuperscript{\rm 1},
        Kai Song,
        Yuzheng Zhuang\textsuperscript{\rm 2},
        Irwin King\textsuperscript{\rm 1}\\
  The Chinese University of Hong Kong\textsuperscript{\rm 1} \\
  Huawei Noah's Ark Lab\textsuperscript{\rm 2}\\
  \texttt{\{yema21, lijj, king\}@cse.cuhk.edu.hk}\\
  \texttt{\{chidafeng1, zhuangyuzheng\}@huawei.com}\\
  \texttt{songkai.neu@gmail.com}
}

\begin{document}
\maketitle
\begin{abstract}
The natural language generation domain has witnessed great success thanks to Transformer models. Although they have achieved state-of-the-art generative quality, they often neglect generative diversity. Prior attempts to tackle this issue suffer from either low model capacity or over-complicated architectures. Some recent methods employ the VAE framework to enhance diversity, but their latent variables fully depend on the input context, restricting exploration of the latent space. In this paper, we introduce VOLTA, a framework that elevates generative diversity by bridging Transformer with VAE via a more effective cross-attention-based connection, departing from conventional embedding concatenation or summation. Additionally, we propose integrating InfoGAN-style latent codes to enable input-independent variability, further diversifying the generation. Moreover, our framework accommodates discrete inputs alongside its existing support for continuous inputs. We perform comprehensive experiments with two types of Transformers on six datasets from three different NLG tasks to show that our approach can significantly improve generative diversity while maintaining generative quality.
\end{abstract}

\section{Introduction}
The rapid advancement of Natural Language Generation (NLG) has been propelled by the remarkable success of Transformer models, including the notable series of GPT models \cite{radford2018improving, radford2019language, brown2020language, DBLP:conf/nips/Ouyang0JAWMZASR22, chatgpt-openai, DBLP:journals/corr/abs-2303-08774}, T5 \cite{DBLP:journals/jmlr/RaffelSRLNMZLL20}, OPT \cite{DBLP:journals/corr/abs-2205-01068}, and LLaMA \cite{DBLP:journals/corr/abs-2302-13971}. While they have demonstrated unparalleled proficiency in autoregressive text generation~\cite{li2020unsupervised,hu2022planet,DBLP:conf/aaai/0007LGKL22,qiu2024clongeval}, they predominantly focus on learning to reassemble text from large corpora with high generative quality. However, the pursuit of generative diversity remains a critical yet underexplored frontier in NLG. Generative diversity is distinct from mere paraphrasing, as it encompasses not only altered syntax but also varied semantics. Early attempts, such as diverse beam search \cite{DBLP:conf/aaai/VijayakumarCSSL18}, have made strides in enhancing diversity by modifying the decoding process. Nonetheless, these methods often fall short in enhancing the model itself, limiting their ability to significantly improve diversity.

\begin{table}[t]
 \centering
 \resizebox{1\columnwidth}{!}{%
    \begin{tabular}{ p{0.1cm} p{7.9cm} }
     \toprule
        \multirow{7}{*}{\rotatebox[origin=c]{90}{\textbf{Context}}} &
        \setlength\fboxsep{1pt}
        Atop the Main Building's gold dome is a \colorbox{red!30}{golden} \colorbox{red!30}{statue of the Virgin Mary}. Immediately in front of the Main Building and facing it, is \colorbox{green!30}{a copper statue} \colorbox{green!30}{of Christ with arms upraised with the legend "Ven-} \colorbox{green!30}{ite Ad Me Omnes"}. Next to the Main Building is the Basilica of the Sacred Heart. Immediately behind the basilica is the \colorbox{orange!50}{Grotto} \mycdots \mycdots \\
        \hline
        \textbf{Q} & What type of statue is on the main building?\\
        \textbf{A} &
            \setlength\fboxsep{1pt}
            \colorbox{red!30}{golden statue of the Virgin Mary}\\
        \hline
        \textbf{Q} & What is the name of the copper statue on the main building?\\
        \textbf{A} &
            \setlength\fboxsep{1pt}
            \colorbox{green!30}{a copper statue of Christ with arms upraised with \mycdots} \\
        \hline
        \textbf{Q} & What is next to the main building?\\
        \textbf{A} &
            \setlength\fboxsep{1pt}
            \colorbox{orange!50}{Grotto}\\
     \bottomrule
    \end{tabular}
 }
 \caption{Examples of generative diversity by VOLTA on the QAG task. Our framework enables the generation of three distinct question-answer pairs.}
 \label{tbl:latent-variable}
\end{table}

\begin{figure}[t]
    \centering
    \includegraphics[width=1\columnwidth,trim={1.8cm 20.5cm 34cm 1.7cm},clip]{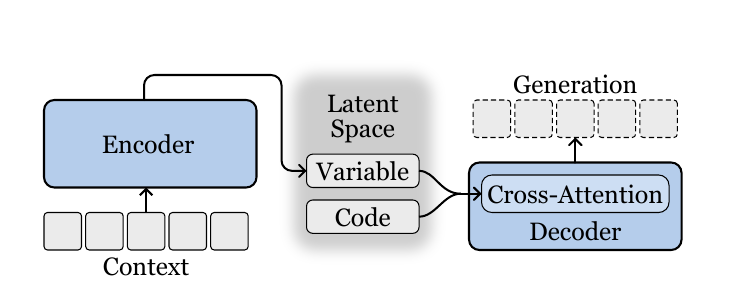}
    \caption{The overview of VOLTA. The encoder encodes the context into VAE latent variables. The variables, augmented with InfoGAN-style \textbf{latent codes}, can be continuous or discrete based on the input type. Subsequently, they are connected to the decoder through the \textbf{cross-attention} mechanism. Leveraging the variability inherent in the latent space, the decoder generates diverse content conditioned on the context.}
    \label{fig:overview}
\end{figure}

Variational Autoencoder (VAE) \cite{DBLP:journals/corr/KingmaW13} offers a framework addressing the low-diversity issue. By encoding inputs into lower-dimensional latent variables, VAE introduces the opportunity to diversify the decoding process: perturbing these latent variables allows generated sentences to deviate from annotated ones, thereby enhancing diversity. However, prior attempts like Info-HCVAE \cite{DBLP:conf/acl/LeeLJKH20}, utilizing LSTM-based VAEs, inherit limitations associated with LSTMs. While Transformers have emerged as the mainstream network, integrating them into the VAE framework poses challenges due to the parallelized self-attention mechanism. More precisely, this complexity arises from inserting a bottleneck layer of latent variables between Transformer layers, as the embeddings of the entire sequence pass through the model simultaneously. Optimus \cite{DBLP:conf/emnlp/LiGLPLZG20}, pioneering the fusion of Transformers with VAEs, adopts BERT \cite{DBLP:conf/naacl/DevlinCLT19} as the VAE encoder and GPT-2 \cite{radford2019language} as the VAE decoder. Subsequent works attempt to improve upon Optimus \cite{DBLP:conf/emnlp/0001HDZJMS22, DBLP:journals/corr/abs-2205-05862, DBLP:conf/emnlp/DengPSC23}, yet they fall short in addressing its three major drawbacks. Firstly, it introduces embedding concatenation and summation to connect latent variables to the decoder, with Optimus performing optimally only upon their combined use. In contrast, our novel cross-attention-based connection proves more effective. Secondly, Optimus's model architecture is overly intricate. It relies on two distinct Transformer models, necessitating two unique tokenizers and extensive pretraining. We streamline this complexity by either employing a shared Transformer decoder as the backbone network or leveraging an encoder-decoder Transformer model. This renders our framework compatible even with Large Language Models (LLMs) such as LLaMA \cite{DBLP:journals/corr/abs-2302-13971} or GPT-4 \cite{DBLP:journals/corr/abs-2303-08774}. Lastly, while Optimus solely handles continuous latent variables, VOLTA expands its scope to cover discrete inputs by encoding them into discrete latent variables, enriching the model's generalizability.

The VAE framework offers increased generative diversity, yet its input-dependent latent variables limit exploration within the latent space, restricting the model's ability to generate a wider array of diverse content. In pursuit of an input-independent approach to vary the generation process, we propose attaching latent codes to VAE latent variables, inspired by InfoGAN \cite{DBLP:conf/nips/ChenCDHSSA16}. Our method employs the Variational Mutual Information Maximization (VMIM) objective to encourage the decoder to autonomously identify distinct semantic features via latent codes. Consequently, this enables more variability in generated content without any reliance on the input. To the best of our knowledge, our work represents the first utilization of latent codes within NLG.

Our framework, dubbed \textbf{VOLTA} (\textbf{\underline{V}}ariati\textbf{\underline{O}}nal Mutua\textbf{\underline{L}} Informa\textbf{\underline{T}}ion Maximizing \textbf{\underline{A}}utoencoder), derives its name from its adherence to the Variational Autoencoder framework and the incorporation of the Variational Mutual Information Maximization objective from InfoGAN. To validate the effectiveness of VOLTA, we benchmark it against state-of-the-art baseline models across six datasets from three representative NLG tasks: language modeling, question-answer generation, and dialog response generation. We also conduct comprehensive ablation studies to examine the impact of the different components of VOLTA.

The main contributions of this paper are:
\begin{itemize}
    \item VOLTA proposes a novel cross-attention mechanism to integrate Transformers with VAEs. It exhibits generalizability to both latent variables (continuous \& discrete) and various Transformer architectures (decoder-only \& encoder-decoder).
    \item To attain input-independent variability, we propose attaching InfoGAN-style latent codes to VAE latent variables.
    \item Comprehensive experimental results on six datasets spanning three distinct NLG tasks validate the efficacy of our model in enhancing generative diversity while upholding quality.
\end{itemize}

\section{Related Work}
In recent years, a multitude of Transformer-based models have emerged, such as the GPT series \cite{radford2018improving, radford2019language, brown2020language, DBLP:conf/nips/Ouyang0JAWMZASR22, chatgpt-openai, DBLP:journals/corr/abs-2303-08774}, T5 \cite{DBLP:journals/jmlr/RaffelSRLNMZLL20}, OPT \cite{DBLP:journals/corr/abs-2205-01068}, and LLaMA \cite{DBLP:journals/corr/abs-2302-13971}. These models are primarily trained to optimize the alignment between generated content and annotations, often prioritizing quality over diversity in the generative process. 

Variational Autoencoders (VAEs) \cite{DBLP:journals/corr/KingmaW13} represent a powerful approach to diverse generation in NLG. They diverge from Autoencoders \cite{hinton2006reducing} by introducing probabilistic latent variables. Originally applied in computer vision, VAEs were later adapted for natural language processing. Early attempts, such as Info-HCVAE \cite{DBLP:conf/acl/LeeLJKH20}, employed LSTMs \cite{6795963} as both encoder and decoder, achieving diversity in question-answer generation (QAG). However, these LSTM-based models suffered from architectural complexities, utilizing separate LSTM modules for encoding and decoding context, questions, and answers. Optimus \cite{DBLP:conf/emnlp/LiGLPLZG20} addressed some of these challenges by using BERT \cite{DBLP:conf/naacl/DevlinCLT19} as the encoder and GPT-2 \cite{radford2019language} as the decoder, surpassing LSTM-based models in VAE language modeling. Subsequent models like VarMAE \cite{DBLP:conf/emnlp/0001HDZJMS22} focused on applying VAEs in language understanding, while RegaVAE \cite{DBLP:conf/emnlp/DengPSC23} attempted augmentation through retrieval methods, and AdaVAE \cite{DBLP:journals/corr/abs-2205-05862} explored the usage of two adaptive GPT-2 models. Our VOLTA model further simplifies the architecture by leveraging a shared backbone network or utilizing an encoder-decoder Transformer model.

In pursuit of more variability and subsequently higher diversity, several methods have employed unique strategies such as special prompt tokens or control phrases. These include SimpleTOD \cite{DBLP:conf/nips/Hosseini-AslMWY20}, CTRL \cite{Keskar2019CTRLAC}, Soloist \cite{DBLP:journals/tacl/PengLLSLG21}, CGRG \cite{DBLP:conf/aaai/WuGBZ0QKGHOD21}, and MEGATRON-CNTRL \cite{DBLP:conf/emnlp/XuPSPFAC20}. \citet{DBLP:conf/iclr/DathathriMLHFMY20} proposed the Plug and Play Language Model, which guides language generation by plugging simple attribute classifiers into existing language models. InfoGAN \cite{DBLP:conf/nips/ChenCDHSSA16} originally controlled image generation using latent codes trained with the Variational Mutual Information Maximization (VMIM) objective. In computer vision, attempts to merge InfoGAN with VAE for controllable generative models have resulted in models like VAE-Info-cGAN \cite{DBLP:conf/gis/XiaoGP20} and InfoVAEGAN \cite{DBLP:conf/icip/YeB21}. However, InfoVAE \cite{DBLP:conf/aaai/ZhaoSE19}, InfoMax-VAE \cite{DBLP:conf/isit/Lotfi-RezaabadV20}, \citet{DBLP:journals/jmlr/Melis0B22}, and VAE-MINE \cite{DBLP:conf/emnlp/QianC19} applied VMIM to VAEs to address the latent variable collapse problem rather than focusing on improving variability. To the best of our knowledge, our model is the first to integrate Transformer models with the VAE and InfoGAN frameworks in Natural Language Generation (NLG). Although we focus on diversity in this paper, other aspects of NLG are also worth exploring in the future \cite{DBLP:conf/cikm/SongZK23, DBLP:conf/nips/SongZK23, DBLP:conf/nips/SongZK23a, DBLP:conf/aaai/MaSHLZK23}, such as multi-modality, bias, and fairness.

\section{Our Method}
Our VOLTA framework is meticulously designed to facilitate diverse generation, leveraging latent variables from the VAE framework \cite{DBLP:journals/corr/KingmaW13} in conjunction with InfoGAN-style latent codes \cite{DBLP:conf/nips/ChenCDHSSA16}. Initially, VOLTA encodes the input into latent variables. Subsequently, by sampling new latent variables, slight alterations in the decoded content can be achieved, promoting greater diversity. Differing from VAE latent variables, InfoGAN-style latent codes operate independently of the input, providing the freedom to explore a broader latent space. This distinct attribute offers an alternative avenue to introduce increased variability within the generated sequences. Figure~\ref{fig:overview} includes an overview of VOLTA.

\subsection{Preliminaries}
In the natural language generation domain, various tasks exist, including language modeling, dialog response generation, and question-answer generation. Generally, NLG aims to generate a new sequence $\bm{x}_{g}=[x_{g,1}, \dots, x_{g,n}]$ based on a provided context sequence $\bm{x}_{c}=[x_{c,1}, \dots, x_{c,m}]$, where each $x$ represents an individual token. The objective is to identify a model $f(\cdot)$ capable of generating an appropriate sequence using the given context: $f(\bm{x}_{c}) \rightarrow \bm{x}_{g}$. In cases like extractive answer generation, the answer is denoted by a pair of integer indices $(s, e)\in \mathbb{N}^2$, indicating the start and end positions of the answer span. Then the answer tokens $\bm{x}_{a} = [x_{c,s}, \cdots, x_{c,e}]$ can be located within the context sequence $\bm{x}_{c}$. It constitutes a part of $\bm{x}_{g}$ unless explicitly specified otherwise.

\subsection{Model Architecture}
\label{sec:volta}
VOLTA adheres to the VAE framework, where the encoder $f_{enc}(\cdot)$ and the decoder $f_{dec}(\cdot)$ are both Transformer models. Unlike Optimus \cite{DBLP:conf/emnlp/LiGLPLZG20}, which utilizes BERT as the encoder and GPT-2 as the decoder, our model offers the simplicity of a shared backbone network between the encoder and decoder. Additionally, VOLTA can adapt encoder-decoder Transformers \cite{DBLP:conf/nips/VaswaniSPUJGKP17} seamlessly into VAE, leveraging their inherent encoder and decoder architecture.

\paragraph{Latent variables from encoder.} The encoder encodes the input sequence into multiple independent continuous or discrete latent variables, selecting the most suitable based on the input type. For instance, dialog responses and questions are aptly represented using continuous latent variables, aligning with their semantic nature. Conversely, discrete latent variables prove advantageous in modeling answer spans, aligning with their positions within the context. Specifically, latent variables can be calculated as follows:
\begin{equation} \label{eq:encoding}
\begin{split}
& \bm{h}_{enc} = f_{enc}(\bm{x}_{c}, \bm{x}_{g}),\\
& \mu_i, \sigma_i = \text{FC}(\bm{h}_{enc}), \quad \bm{\pi}_{j} = \text{FC}(\bm{h}_{enc}),\\
& z_{g,i} \sim \mathcal{N}(\mu_i, \sigma_i^{2}), \hspace{20pt} z_{a,j} \sim \text{Cat}(\bm{\pi}_{j}),
\end{split}
\end{equation}
where $\text{FC}(\cdot)$ is a single fully-connected layer and each instance has a distinct set of learnable parameters; indexing is omitted for simplicity; $\mathcal{N}(\cdot)$ is the Gaussian distribution with parameters $\mu_i$ and $\sigma_i$; $\text{Cat}(\cdot)$ is the categorical distribution whose parameters $\bm{\pi}_{j}$ represent the event probabilities of $k$ categories. Back-propagation through the latent variables is achieved using the Gaussian distribution reparameterization trick \cite{DBLP:conf/fair2/WolpeW19} for $\bm{z}_g=[z_{g,1}, z_{g,2}, z_{g,3}, \mycdots]$ and Gumbel-Softmax \cite{DBLP:conf/iclr/MaddisonMT17, DBLP:conf/iclr/JangGP17} reparameterization for $\bm{z}_a=[z_{a,1}, z_{a,2}, z_{a,3}, \mycdots]$.

\paragraph{Latent codes.} Supplementing the VAE latent variables, we incorporate InfoGAN-style latent codes \cite{DBLP:conf/nips/ChenCDHSSA16} to infuse the model with input-independent variability. These latent codes come in two types: continuous latent codes, which can conform to either uniform distribution or Gaussian distribution, and discrete latent codes, which also adhere to categorical distribution:
\begin{equation}
\begin{split}
\bm{c}_{g} &= [c_{g,1}, c_{g,2}, c_{g,3}, \mycdots], \hspace{5pt} c_{g,i} \sim \text{Uni}(-1, 1), \\
\bm{c}_{a} &= [c_{a,1}, c_{a,2}, c_{a,3}, \mycdots], \hspace{5pt} c_{a,j} \sim \text{Cat}(\bm{\rho}), 
\end{split}
\end{equation}
where $\text{Uni}(\cdot)$ is the uniform distribution; the categorical distribution has parameters $\bm{\rho}=\frac{1}{k}\mathbf{1}$, which uses the same number of categories $k$ as the discrete latent variables $\bm{z}_a$. This compatibility is necessary as they will be concatenated together.

\paragraph{Cross-attention-based latent-space connection.} Optimus \cite{DBLP:conf/emnlp/LiGLPLZG20} uses two channels to connect latent variables to the decoder: the ``embedding'' channel involves a fully-connected layer to obtain a latent embedding, which is subsequently added to word embeddings. Meanwhile, the ``memory'' channel generates latent embeddings for each Transformer block within the decoder. These latent embeddings are then concatenated with decoder hidden states as past information. The optimal performance is attained when both channels are utilized, 
 albeit complicating the architecture.

In Transformers \cite{DBLP:conf/nips/VaswaniSPUJGKP17}, the attention mechanism can take the form of self-attention or cross-attention. We introduce a unified and notably more effective cross-attention-based connection between the latent space and the decoder:
\begin{align}
\begin{split}
K_{\text{latent}} &= \text{FC}([\bm{z} , \bm{c}]), \\
V_{\text{latent}} &= \text{FC}([\bm{z} , \bm{c}]), \\
\text{Attention} & (Q, K_{\text{latent}}, V_{\text{latent}}) \\
&= \text{softmax}(\frac{QK_{\text{latent}}^\mathsf{T}}{\sqrt{d_k}})V_{\text{latent}}.
\end{split}
\end{align}
We facilitate the transmission of latent space information into the decoder using $K_{\text{latent}}$ and $V_{\text{latent}}$, queried by the decoder via $Q$. In cases where the Transformer model lacks pretrained weights for cross-attention layers, such as decoder-only Transformers, we retain Optimus's connection method. However, we streamline it by incorporating a shared backbone for both the encoder and decoder.

\begin{table*}[ht]
 \centering
 \resizebox{1\textwidth}{!}{%
    \begin{tabular}{ l  l  c  c  c  c  c  c  c  c }
     \toprule
        & \multicolumn{2}{c}{\textbf{Specifications}} &  \multicolumn{2}{c}{\textbf{Quality}}  & \multicolumn{5}{c}{\textbf{Diversity}}  \\

        \textbf{Model} & Type & Para. & EM & F1 & Dist-1 & Dist-2 & Dist-3 & Dist-4 & S-BL $\downarrow$\\
        \cmidrule(lr){1-1} \cmidrule(lr){2-3} \cmidrule(lr){4-5} \cmidrule(lr){6-10}

        GPT-2  & TFM-Dec & $124$M &
            $56.28$ & $67.86$ & $8.23$ & $38.63$ & $62.58$ & $75.42$ & $32.09$ \\

        \cmidrule(lr){1-1} \cmidrule(lr){2-3} \cmidrule(lr){4-5} \cmidrule(lr){6-10}
        BART & TFM-Enc-Dec & $139$M &
            $58.03$ & $69.99$ & $8.08$ & $38.49$ & $62.34$ & $74.91$ & $32.66$ \\

        \cmidrule(lr){1-1} \cmidrule(lr){2-3} \cmidrule(lr){4-5} \cmidrule(lr){6-10}
        T5  & TFM-Enc-Dec & $222$M &
            $59.76$ & $71.98$ & $8.18$ & $40.78$ & $65.52$ & $77.20$ & $30.51$ \\

        \cmidrule(lr){1-1} \cmidrule(lr){2-3} \cmidrule(lr){4-5} \cmidrule(lr){6-10}
        OPT  & TFM-Dec & $331$M &
            $58.57$ & $70.40$ & $7.88$ & $38.51$ & $63.80$ & $76.55$ & $29.97$ \\

        \cmidrule(lr){1-1} \cmidrule(lr){2-3} \cmidrule(lr){4-5} \cmidrule(lr){6-10}
        HCVAE & VAE w/ LSTM & $158$M &
            $61.81$ & $73.68$ & $7.00$ & $33.47$ & $57.24$ & $71.68$ & $32.66$  \\

        \cmidrule(lr){1-1} \cmidrule(lr){2-3} \cmidrule(lr){4-5} \cmidrule(lr){6-10}
        Optimus & VAE w/ TFM & $233$M &
            $58.05$ & $69.55$ & $8.05$ & $40.27$ & $66.63$ & $79.88$ & $29.28$ \\

        \cmidrule(lr){1-1} \cmidrule(lr){2-3} \cmidrule(lr){4-5} \cmidrule(lr){6-10}
        \textbf{VOLTA} & VAE w/ TFM & $124$M &
            $\boldsymbol{65.56}$ & $\boldsymbol{77.31}$ & $\boldsymbol{8.32}$ & $\boldsymbol{40.84}$ & $\boldsymbol{68.05}$ & $\boldsymbol{82.64}$ & $\boldsymbol{28.34}$ \\
     \bottomrule
    \end{tabular}
 }
 \caption{Performance comparison on question-answer generation. \textit{Abbreviations}: ``HCVAE'': Info-HCVAE; ``Para.'': Parameter Count; ``Dist-k'': Distinct-k; ``S-BL'': Self-BLEU; ``TFM'': Transformer; ``Enc'', ``Dec'': Encoder, Decoder; ``$\downarrow$'' means lower is better. }
 \label{tbl:comparison}
\end{table*}

\paragraph{Generation.} VOLTA is trained in the typical autoregressive manner to predict subsequent tokens by considering the preceding tokens:
\begin{align}  \label{eq:qtn-gen} 
\begin{split}
\bm{h}_{g,t} &= f_{dec}(\bm{x}_{c}, \bm{x}_{g,<t}, [\bm{z}_g , \bm{c}_g]), \\
p(\bm{x}_{g})
&= \prod_{t=1}^{n} p(x_{g,t} \mid \bm{x}_{c}, \bm{x}_{g,<t}, [\bm{z}_g , \bm{c}_g])  \\ 
&= \prod_{t=1}^{n} \text{softmax}(\text{FC}(\bm{h}_{g,t})),
\end{split}
\end{align}
where $\bm{x}_{g,<t}$ means the first $t-1$ tokens in $\bm{x}_{g}$.

The process for generating discrete data follows a similar approach but involves a distinct prediction head. Specifically, in the scenario of answer generation:
\begin{equation} \label{eq:ans-gen}
\begin{split}
& \bm{h}_a = f_{dec}(\bm{x}_{c}, [\bm{z}_a, \bm{c}_a]),\\
& p(s) = \text{softmax}(\text{FC}(\bm{h}_{a,1:m})), \\ 
& p(e) = \text{softmax}(\text{FC}(\bm{h}_{a,1:m})),\\
& s = \argmax_{s\in \{1, \mycdots, m\}} p(s),  \\
& e = \argmax_{e\in \{1, \mycdots, m\}} p(e),\\
& \bm{x}_{a} = [x_{c,s}, \cdots, x_{c,e}], \\
\end{split}
\end{equation}
where $\bm{h}_a$ denotes the hidden states obtained from the decoder; the subscript $1:m$ means slicing an array from index $1$ to $m$, which corresponds to the context tokens. This results in a generated answer $\bm{x}_{a}$, where $s$ denotes the starting index and $e$ denotes the ending index.

\subsection{Training Objectives}
\label{sec:objectives}

Since the marginal likelihood $p(\bm{x})$ is intractable to compute, we approximate the true posterior $p(z\mid\bm{x})$ with $q(z \mid \bm{x})$ based on our encoder $f_{enc}(\cdot)$. Following the standard VAE formulation, we define the evidence lower bound (ELBO) as $\text{ELBO} = - \mathcal{L}_{\text{AE}}(\bm{x}) - \mathcal{L}_{\text{REG}}(\bm{z}) $. Here, $\mathcal{L}_{\text{AE}}$ stands for Autoencoder (AE) reconstruction loss and $\mathcal{L}_{\text{REG}}$ represents $D_{\text{KL}}(q(z \mid \bm{x}) \parallel p(z))$ for regularization.

\paragraph{Latent variable regularization loss.} The KL divergence for regularizing the continuous or discrete latent variable is:
\begin{align} \label{eq:KL-divergence}
& \mathcal{L}_{\text{REG}}(z_g) = \log \frac{\sigma'}{\sigma} + \frac{\sigma^2+(\mu - \mu')^2}{2\sigma'^2}-\frac{1}{2},  \notag \\
& \mathcal{L}_{\text{REG}}(z_a) = \sum_{i=1}^{k} \pi_{i} \log \frac{\pi_{i}}{\pi'_{i}},
\end{align}
where $\mu, \sigma, \bm{\pi}$ follows Eq.~(\ref{eq:encoding}); we assume that the priors $p(z_{g})$ and $p(z_{a})$ follow $\mathcal{N}(\mu', \sigma'^{2})$ and $\text{Cat}(\bm{\pi'})$, respectively. In practice, $\mu'$, $\sigma'$ and $\bm{\pi'}$ can be obtained by encoding only the context $\bm{x}_c$. The total $\mathcal{L}_{\text{REG}}$ is the mean over the latent variables. The derivations are in Appendix~\ref{appendix:optimus}, \ref{appendix:hcvae}.

\paragraph{Latent code VMIM loss.} To prevent the model from ignoring the latent codes, we encourage it to recover the latent codes in the generation phase by optimizing the Variational Mutual Information Maximization (VMIM) objective \cite{DBLP:conf/nips/ChenCDHSSA16}:
\begin{align}
\begin{split}
& I(c; f_{dec}(\bm{x}, [\bm{z} , c])) \\
={}& H(c) + \mathbb{E}_{\bm{x'}} \Big[
    \begin{aligned}[t] 
        &  D_\text{KL}\big(p(c' \mid \bm{x'}) \parallel q(c' \mid \bm{x'})\big) \\ 
        & + \mathbb{E}_{c'}\big[\log q(c' \mid \bm{x'})\big] \Big] \\
    \end{aligned}    \\ 
\geq & H(c) + \mathbb{E}_{\bm{x'}} \Big[ \mathbb{E}_{c'}\big[\log q(c'\mid \bm{x'}) \big] \Big] \\
\triangleq & H(c) - \mathcal{L}_{\text{VMIM}}(c) , 
\end{split}
\end{align}
where $\bm{x'}\sim f_{dec}(\bm{x}, [\bm{z} , c])$; $c'\sim p(c \mid \bm{x'})$ is the recovered latent code. Because the posterior $p(c \mid \bm{x'})$ is difficult to obtain, an auxiliary distribution $q(c \mid \bm{x'})$ based on $f_{dec}(\cdot)$ is added to approximate it. The entropy $H(c)$ is a constant and thus excluded from $\mathcal{L}_{\text{VMIM}}(c)$. The derivation of this objective is included in Appendix~\ref{appendix:infogan}.

In our model, a fully-connected layer is added to the decoder for recovering each latent code $c$:
\begin{align} \label{eq:VMIM}
\begin{split}
\theta &= \text{FC}(f_{dec}(\bm{x}, [\bm{z} , c])),\\
\mathcal{L}_{\text{VMIM}}(c) &= -\log q(c'; \theta),\\
\end{split}
\end{align}
where the parameter $\theta$ depends on the distribution type of the corresponding latent code $c$. The total VMIM loss is the mean over the latent codes.

\paragraph{Overall objective.} By Eq.~(\ref{eq:KL-divergence})(\ref{eq:VMIM}), the overall loss:
\begin{equation}
 \mathcal{L}(\bm{x})=\mathcal{L}_{\text{AE}}(\bm{x}) + \beta \mathcal{L}_{\text{REG}}(\bm{z}) + \gamma \mathcal{L}_{\text{VMIM}}(\bm{c}),
\end{equation}
where $\beta$, $\gamma$ denote the coefficients used to adjust the loss weights; the Autoencoder reconstruction loss $\mathcal{L}_{\text{AE}}(\bm{x})$ corresponds to the standard cross-entropy loss employed for language modeling.

\begin{table*}[ht]
 \centering
 \resizebox{1\textwidth}{!}{%
    \begin{tabular}{l  c  c  c  c  c  c  c  c  c  c  c  c }
     \toprule
        \textbf{Dataset} & \multicolumn{3}{c}{\texttt{PTB}} & \multicolumn{3}{c}{\texttt{YELP}} & \multicolumn{3}{c}{\texttt{YAHOO}} & \multicolumn{3}{c}{\texttt{SNLI}} \\

        \textbf{Model} & PPL $\downarrow$ & MI & AU & PPL $\downarrow$ & MI & AU & PPL $\downarrow$ & MI & AU & PPL $\downarrow$ & MI & AU \\
        \cmidrule(lr){1-1} \cmidrule(lr){2-4} \cmidrule(lr){5-7} \cmidrule(lr){8-10} \cmidrule(lr){11-13}

        M. A.   &101.40 & 0.00 & 0  & 40.39 & 0.13 & 1  & 61.21 & 0.00 & 0  & 21.50 & 1.45 & 2  \\
        \cmidrule(lr){1-1} \cmidrule(lr){2-4} \cmidrule(lr){5-7} \cmidrule(lr){8-10} \cmidrule(lr){11-13}

        C. A.    &108.81 & 1.27 & 5  & -     & -    & -  & 66.93 & 2.77 & 4  & 23.67 & 3.60 & 5  \\
        \cmidrule(lr){1-1} \cmidrule(lr){2-4} \cmidrule(lr){5-7} \cmidrule(lr){8-10} \cmidrule(lr){11-13}

        SA-VAE   & -     & -    & -  & -     & 1.70 & 8  & 60.40 & 2.70 & 10 & -     & -    & -  \\
        \cmidrule(lr){1-1} \cmidrule(lr){2-4} \cmidrule(lr){5-7} \cmidrule(lr){8-10} \cmidrule(lr){11-13}

        Aggressive & 99.83 & 0.83 & 4  & 39.84 & 2.16 & 12 & 59.77 & 2.90 & 19 & \textbf{21.16} & 1.38 & 5  \\
        \cmidrule(lr){1-1} \cmidrule(lr){2-4} \cmidrule(lr){5-7} \cmidrule(lr){8-10} \cmidrule(lr){11-13}

        AE-BP     & 96.86 & 5.31 & 32 & 47.97 & 7.89 & 32 & 59.28 & 8.08 & 32 & 21.64 & 7.71 & 32 \\
        \cmidrule(lr){1-1} \cmidrule(lr){2-4} \cmidrule(lr){5-7} \cmidrule(lr){8-10} \cmidrule(lr){11-13}

        Optimus & 51.39 & 0.02 & 0  & 27.63 & 0.02 & 0  & 29.35 & 0.04 & 0  & 66.58 & 9.20 & 32 \\

        \cmidrule(lr){1-1} \cmidrule(lr){2-4} \cmidrule(lr){5-7} \cmidrule(lr){8-10} \cmidrule(lr){11-13}
        \textbf{VOLTA} & \textbf{45.29} & \textbf{8.17} & 32 & \textbf{14.14} & \textbf{9.00} & 32 & \textbf{14.82} & \textbf{9.02} & 32 & 25.69 & \textbf{9.24} & 32  \\

     \bottomrule
    \end{tabular}
 }
 \caption{Performance comparison on language modeling tasks. Baseline results are obtained from \citet{DBLP:conf/emnlp/LiGLPLZG20}, except for Optimus, which omits second-stage pretraining for fair comparison. The maximum achievable AU is 32.}
 \label{tbl:lm}
\end{table*}

\begin{table}[t]
 \centering
 \resizebox{1\columnwidth}{!}{%
    \begin{tabular}{l  c  c  c  }
     \toprule
        & \textbf{Quality} & \textbf{Diversity} & \textbf{Overall} \\
        \textbf{Model} & Precision & Recall & F1 \\

        \hline
        Seq2Seq & 0.232 & 0.232 & 0.232 \\
        \hline
        SeqGAN & 0.270 & 0.270 & 0.270 \\
        \hline
        CVAE & 0.222 & 0.265 & 0.242 \\
        \hline
        VHRED & 0.341 & 0.278 & 0.306 \\
        \hline
        VHCR & 0.271 & 0.260 & 0.265 \\
        \hline
        WAE & 0.266 & 0.289 & 0.277 \\
        \hline
        iVAE$_{\text{MI}}$ & 0.239 & 0.355 & 0.285 \\
        \hline
        T5 & 0.321 & 0.321 & 0.321 \\
        \hline
        Optimus & 0.313 & 0.362 & 0.336 \\
        \hline
        \textbf{VOLTA} & \textbf{0.373} & \textbf{0.401} & \textbf{0.387} \\

     \bottomrule
    \end{tabular}
 }
 \caption{Performance comparison on dialog response generation. Baseline results are from \citet{DBLP:conf/emnlp/LiGLPLZG20}, except for T5.}
 \label{tbl:dialog}
\end{table}

\section{Experiments}

\subsection{Tasks and Datasets}
We evaluate VOLTA against various baselines across six datasets, spanning three distinct NLG tasks:
\begin{itemize}
    \item \textbf{Question-answer generation} (QAG): we employ the SQuAD dataset \cite{DBLP:conf/emnlp/RajpurkarZLL16}, with approximately 100K question-answer pairs where the answers are extractive;
    \item \textbf{Language modeling}: we evaluate on four LM datasets: Penn Treebank (PTB) \cite{DBLP:journals/coling/MarcusSM94}, SNLI \cite{DBLP:conf/emnlp/BowmanAPM15}, YELP, and YAHOO \cite{DBLP:conf/icml/YangHSB17, DBLP:conf/iclr/HeSNB19};
    \item \textbf{Dialog response generation}: we utilize the DailyDialog dataset \cite{DBLP:conf/ijcnlp/LiSSLCN17}, comprising approximately 13K multi-turn conversations, averaging eight turns per dialog.
\end{itemize}

\subsection{Implementation Details}
We conduct experiments using two Transformer model variants: the decoder-only Transformer, leveraging the GPT-2 base model \cite{radford2019language}, and the encoder-decoder Transformer, utilizing the T5 base model \cite{DBLP:journals/jmlr/RaffelSRLNMZLL20}. With the decoder-only Transformer, since it comprises solely Transformer decoder blocks, we employ it as the shared backbone for both the encoder and decoder within VOLTA. In contrast, the encoder-decoder Transformer features distinct Transformer encoder and decoder, aligning conveniently with the VOLTA encoder and decoder structures. Throughout our experiments, all Transformer-based models load pretrained checkpoints from Hugging Face \footnote{\url{https://huggingface.co/}}, undergoing fine-tuning exclusively on the respective datasets. Unlike the approach in Optimus \cite{DBLP:conf/emnlp/LiGLPLZG20}, no secondary-stage pretraining is executed.

Our model utilizes a default configuration comprising 32 Gaussian latent variables, along with 4 uniform latent codes. For extractive answers, we utilize 20 categorical latent variables and 5 categorical latent codes, all comprising 10 categories, as shown in Table~\ref{tbl:ablation}. Training is performed over 10 epochs with a learning rate set to $5\times 10^{-5}$. To address the KL vanishing issue \cite{DBLP:conf/conll/BowmanVVDJB16}, we employ a linear annealing schedule for $\beta$ \cite{DBLP:conf/emnlp/LiGLPLZG20}. This includes an initial increasing phase covering the first $25\%$ of training, ascending from 0 to a maximum value of 0.1 \cite{DBLP:conf/acl/LeeLJKH20}. Additionally, we set $\lambda = 1.0$ in the KL thresholding scheme \cite{DBLP:conf/emnlp/LiHNBY19} for language modeling. We conduct the experiments on four TITAN V GPUs.

\begin{table*}[t]
 \centering
 \resizebox{1\textwidth}{!}{%
    \begin{tabular}{c c c c c c c c c c c c c c }
     \toprule
        & \multicolumn{5}{c}{\textbf{Configuration}} &  \multicolumn{2}{c}{\textbf{Quality}}  & \multicolumn{5}{c}{\textbf{Diversity}}  \\
        Row & $\bm{z}_g$ & $\bm{z}_a$ & Var. & $\bm{c}_g$ & $\bm{c}_a$ & EM & F1 & Dist-1 & Dist-2 & Dist-3 & Dist-4 & S-BL $\downarrow$\\

        \cmidrule(lr){1-1} \cmidrule(lr){2-6} \cmidrule(lr){7-8} \cmidrule(lr){9-13}

         \texttt{DFLT} & 32 & 20 & \cmark &  Rnd & Rnd &
            $\boldsymbol{65.56}$ & $\boldsymbol{77.31}$ & $8.32$ & $\boldsymbol{40.84}$ & $\boldsymbol{68.05}$ & $\boldsymbol{82.64}$ & $\boldsymbol{28.34}$ \\

                \cmidrule(lr){1-1} \cmidrule(lr){2-6} \cmidrule(lr){7-8} \cmidrule(lr){9-13}
        \texttt{A} & \textcolor{RedOrange}{16} & 20 & \cmark & Rnd & Rnd &
            $63.00$ & $75.45$ & $8.18$ & $38.73$ & $63.54$ & $76.94$ & $34.81$ \\

                \cmidrule(lr){1-1} \cmidrule(lr){2-6} \cmidrule(lr){7-8} \cmidrule(lr){9-13}
        \texttt{B} & \textcolor{RedOrange}{64} & 20 & \cmark & Rnd & Rnd &
            $64.82$ & $76.21$ & $8.11$ & $38.28$ & $63.38$ & $76.98$ & $34.63$ \\

                \cmidrule(lr){1-1} \cmidrule(lr){2-6} \cmidrule(lr){7-8} \cmidrule(lr){9-13}
        \texttt{C} & 32 & \textcolor{RedOrange}{10} & \cmark & Rnd & Rnd &
            $61.73$ & $74.32$ & $\boldsymbol{8.37}$ & $39.66$ & $65.77$ & $80.36$ & $30.75$ \\

                \cmidrule(lr){1-1} \cmidrule(lr){2-6} \cmidrule(lr){7-8} \cmidrule(lr){9-13}
        \texttt{D} & 32 & \textcolor{RedOrange}{40} & \cmark & Rnd & Rnd &
            $62.70$ & $75.11$ & $\boldsymbol{8.37}$ & $39.80$ & $66.08$ & $80.65$ & $30.50$ \\

                \cmidrule(lr){1-1} \cmidrule(lr){2-6} \cmidrule(lr){7-8} \cmidrule(lr){9-13}
        \texttt{E} & 32 & 20 & \textcolor{RedOrange}{\xmark} & Rnd & Rnd &
            $45.72$ & $57.65$ & $5.12$ & $20.39$ & $31.53$ & $37.75$ & $76.04$ \\

                \cmidrule(lr){1-1} \cmidrule(lr){2-6} \cmidrule(lr){7-8} \cmidrule(lr){9-13}
        \texttt{F} & 32 & 20 & \textcolor{RedOrange}{\xmark} & \textcolor{RedOrange}{Fix} & Rnd &
            $46.71$ & $58.49$ & $4.44$ & $16.40$ & $24.51$ & $28.77$ & $84.19$ \\
                \cmidrule(lr){1-1} \cmidrule(lr){2-6} \cmidrule(lr){7-8} \cmidrule(lr){9-13}
        \texttt{G} & 32 & 20 & \textcolor{RedOrange}{\xmark} & Rnd & \textcolor{RedOrange}{Fix} &
            $46.37$ & $58.37$ & $4.50$ & $16.49$ & $24.45$ & $28.59$ & $84.66$ \\

     \bottomrule
    \end{tabular}
 }
 \caption{Ablation study of VOLTA's latent space on the QAG task. The orange text highlights differences from the default configuration in row \texttt{DFLT}. \textit{Abbreviations}: ``Var.'': variational(\cmark) $/$ deterministic(\xmark) latent variables; ``Rnd$/$Fix'': random $/$ fixed latent codes.}
 \label{tbl:ablation}
\end{table*}

\begin{table*}
 \centering
    \begin{tabular}{ p{0.5cm} p{14.5cm} }
     \toprule
        \multirow{3}{*}{\rotatebox[origin=c]{90}{\textbf{Context}}} &
        \setlength\fboxsep{1pt}
        The university is the major seat of the Congregation of Holy Cross (albeit not its official headquarters, which are in Rome). Its main seminary, Moreau Seminary, is located on the campus across St. Joseph lake from the Main Building \mycdots\mycdots
        \\
        \hline
        \textbf{Q1} & What catholic denomination is the university of new haven located in?\\
        \textbf{Q2} & What is the main campus of moreau seminary?\\
        \textbf{Q3} & What religious institution is located on the campus of moreau seminary?\\
        \textbf{Q4} & What former retreat center is located near the grotto?\\
        \textbf{Q5} & What religious denomination does the moreau seminary belong to?\\
        \textbf{Q6} & What is the oldest building on campus?\\
        \textbf{Q7} & What is the main seminary in the university of kansas?\\
        \textbf{Q8} & What is the main seminary of the college?\\
        \textbf{Q9} & What retreat center is located near the grotto?\\
     \bottomrule
    \end{tabular}
 \caption{An example of latent variable interpolation.}
 \label{tbl:latent_var_interpolation}
\end{table*}

\subsection{Metrics}
While our focus lies in achieving diverse NLG, maintaining generative quality is paramount, as completely random sentences might achieve perfect diversity scores but lack meaningful content.

\paragraph{Generative quality.} In dialog response generation, we evaluate generative quality using BLEU-precision \cite{DBLP:conf/acl/PapineniRWZ02}. For language modeling, we measure perplexity (PPL) and mutual information (MI) to assess quality. In question-answer generation, direct measurement against SQuAD reference sentences is not feasible because higher diversity in its nature means shifting the generated content away from these references. Instead, \citet{DBLP:conf/emnlp/ZhangB19} proposed \underline{Q}uestion-\underline{A}nswering-based \underline{E}valuation (QAE), including three main steps: (a) use the QAG model to generate question-answer pairs for raw Wikipedia entries; (b) train a separate question-answering model on the generated QA pairs; (c) evaluate the QA model's performance on the SQuAD development set, using exact match (EM) and F1 metrics \cite{DBLP:conf/emnlp/RajpurkarZLL16, DBLP:conf/acl/RajpurkarJL18}. Poor performance in step (c) reflects low quality of the generated QA pairs, indirectly assessing the QAG model's quality. BERT \cite{DBLP:conf/naacl/DevlinCLT19} serves as the QA model in (b), and we utilize a QAMI loss to enhance QA pair relevance, akin to Info-HCVAE \cite{DBLP:conf/acl/LeeLJKH20}.

\begin{table*}
 \centering
    \begin{tabular}{ p{0.5cm} p{14.5cm} }
     \toprule
        \multirow{3}{*}{\rotatebox[origin=c]{90}{\textbf{Context}}} &
        Holy Cross Father \textbf{John Francis O'Hara} was elected vice-president in 1933 and president of Notre Dame in 1934. During his tenure at Notre Dame, he brought numerous refugee intellectuals to campus; \mycdots \mycdots
        \\
        \hline
        \textbf{Q1} & $c_g=-.8$
            What was O'Hara's first name?\\
        \textbf{Q2} & $c_g=-.6$
            Who was elected vice president in 1933?\\
        \textbf{Q3} & $c_g=-.0$
            What was O'Hara's title prior to becoming vice president?\\
        \textbf{Q4} & $c_g=+.4$
            What was O'Hara's first title?\\
        \hline
        \textbf{A} & John Francis O'Hara\\
     \bottomrule
    \end{tabular}

  \vspace{0.1cm}
    \begin{tabular}{ p{0.5cm} p{14.5cm} }
     \toprule
        \multirow{3}{*}{\rotatebox[origin=c]{90}{\textbf{Context}}} &
        During his 13 years the Irish won \colorbox{red!30}{three} national championships, had \colorbox{green!30}{five} undefeated seasons, won the Rose Bowl in \colorbox{orange!30}{1925}, and produced players such as George Gipp and the "Four Horsemen". \mycdots \mycdots
        \\
        \hline
        \multicolumn{2}{p{15cm}}{
                \textbf{A1}  $c_a=0$  \colorbox{green!30}{five} \hspace{3cm}
                \textbf{A2}  $c_a=3$  \colorbox{orange!30}{1925} \hspace{3cm}
                \textbf{A3}  $c_a=7$  \colorbox{red!30}{three}
                }        \\
     \bottomrule
    \end{tabular}
 \caption{Continuous ($c_g$) $/$ Discrete  ($c_a$) latent code for varying question $/$ answer generation. }
 \label{tbl:latent-code}
\end{table*}

\paragraph{Generative diversity.} In dialog response generation, we assess diversity using BLEU-recall \cite{DBLP:conf/acl/PapineniRWZ02} as the diversity measurement. In language modeling, we analyze the impact of VAE latent variables by tracking the number of active units (AU). Quantitatively measuring diversity in generated questions involves two metrics: Distinct-k \cite{DBLP:conf/naacl/LiGBGD16} and Self-BLEU \cite{DBLP:conf/sigir/ZhuLZGZWY18}. Distinct-k calculates the ratio of distinct k-grams to the total number of generated words. Self-BLEU computes the average BLEU score \cite{DBLP:conf/acl/PapineniRWZ02} for each sentence against all others, aiming for dissimilarity among generated sentences. We generate five QA pairs for each context.

\subsection{Question-Answer Generation}
\label{sec:qag}

We compare VOLTA with several state-of-the-art baselines on the question-answer generation task, as summarized in Table~\ref{tbl:comparison}. We base it on GPT-2 to aim for the minimal model size, showcasing the efficiency of the VOLTA framework. The VAE components in VOLTA add a mere 0.46M parameters. The first four baseline models—GPT-2  \cite{radford2019language}, BART \cite{DBLP:conf/acl/LewisLGGMLSZ20}, T5 \cite{DBLP:journals/jmlr/RaffelSRLNMZLL20} and OPT \cite{DBLP:journals/corr/abs-2205-01068}—all rely on regular Transformer architectures, lacking the variational aspects found in VAE and thereby demonstrating lower generative diversity. Although Info-HCVAE \cite{DBLP:conf/acl/LeeLJKH20} used VAE, it inherited LSTM's limitations. We therefore also adapt the Transformer-based VAE, Optimus \cite{DBLP:conf/emnlp/LiGLPLZG20}, to question generation. Our VOLTA framework harnesses Transformer models' high capacity alongside the variability inherent in VAE and InfoGAN. It stands out by achieving superior diversity over all baselines while maintaining a relatively small model size.

\subsection{Language Modeling}
In language modeling, we employ T5-based VOLTA, comparing it with prior VAE approaches: M. A. \cite{DBLP:conf/conll/BowmanVVDJB16}, C. A. \cite{DBLP:conf/naacl/FuLLGCC19}, SA-VAE \cite{DBLP:conf/icml/KimWMSR18}, Aggressive Training \cite{DBLP:conf/iclr/HeSNB19}, AE-BP \cite{DBLP:conf/emnlp/LiHNBY19}, and the Transformer-based VAE model, Optimus \cite{DBLP:conf/emnlp/LiGLPLZG20}. Our findings reveal that when solely fine-tuned on LM datasets without prior extensive second-stage pretraining on large-scale datasets, the latent variables of the Optimus model \cite{DBLP:conf/emnlp/LiGLPLZG20} collapsed. The reason behind this could lie in Optimus employs two separate latent-space connection methods, which are challenging to optimize. On the contrary, VOLTA's unified cross-attention-based approach proves notably more stable.

\subsection{Dialog Response Generation}
We compare VOLTA with Optimus \cite{DBLP:conf/emnlp/LiGLPLZG20}, the current state-of-the-art model, and several other baselines: Seq2Seq \cite{DBLP:conf/aaai/SerbanSBCP16}, SeqGAN \cite{DBLP:conf/emnlp/LiMSJRJ17}, CVAE \cite{DBLP:conf/acl/ZhaoZE17}, VHRED \cite{DBLP:conf/aaai/SerbanSLCPCB17}, VHCR \cite{DBLP:conf/nips/SubramanianRSTC18}, WAE \cite{DBLP:conf/iclr/GuCHK19}, iVAE$_{\text{MI}}$ \cite{DBLP:conf/emnlp/FangLGDC19}. VOLTA is based on T5 \cite{DBLP:journals/jmlr/RaffelSRLNMZLL20} for dialog response generation, and we include T5 as a baseline to assess the impact of the VOLTA framework. We maintain VOLTA's generation process without incorporating a joint latent space and fusion regularization for history and response \cite{DBLP:conf/naacl/GaoLZBGGD19}, enabling a more general approach compared to Optimus.

\begin{figure*}
    \centering
    \includegraphics[width=1\columnwidth,trim={5cm 2cm 4cm 1cm},clip]{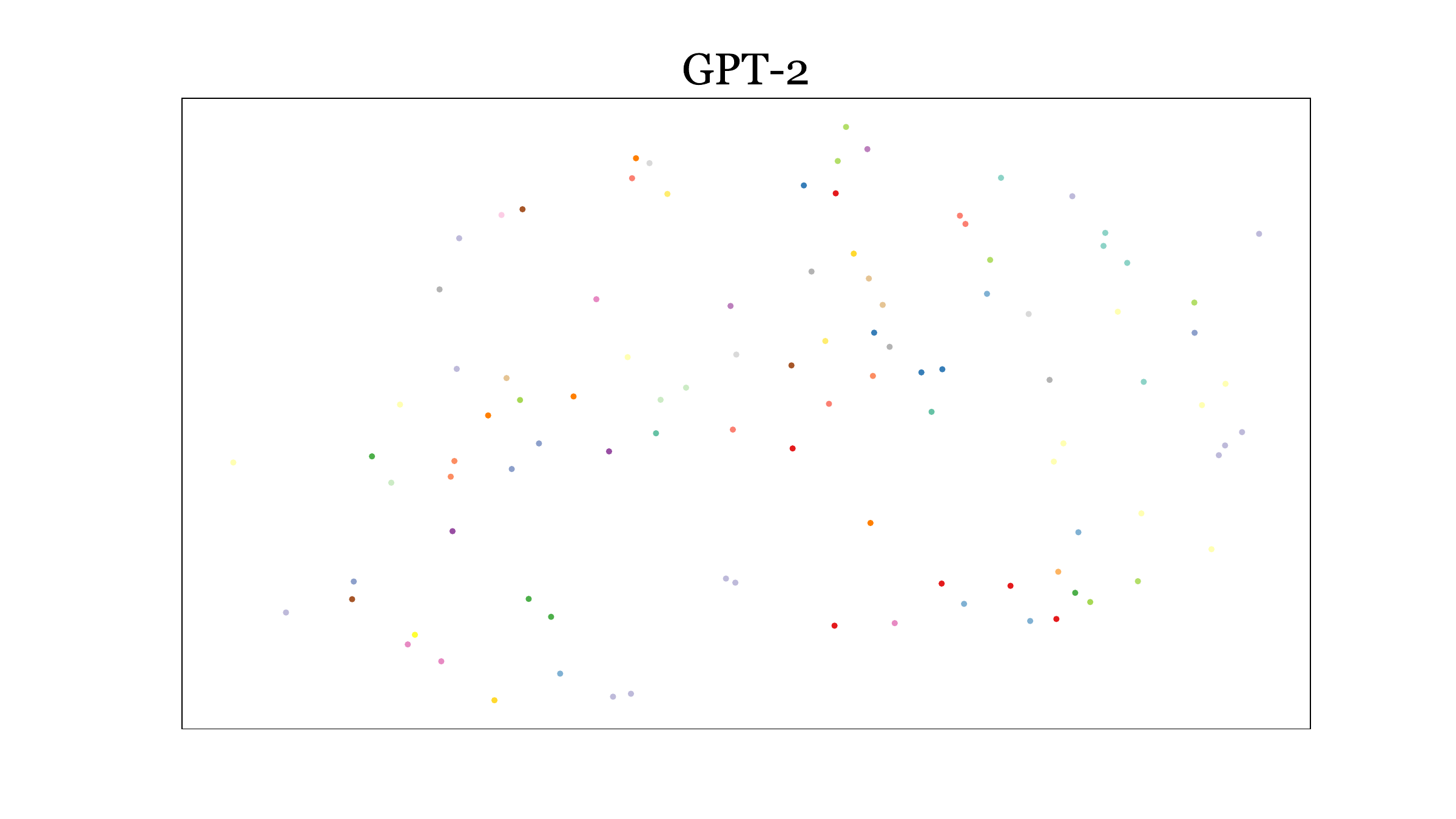}
    \includegraphics[width=1\columnwidth,trim={5cm 2cm 4cm 1cm},clip]{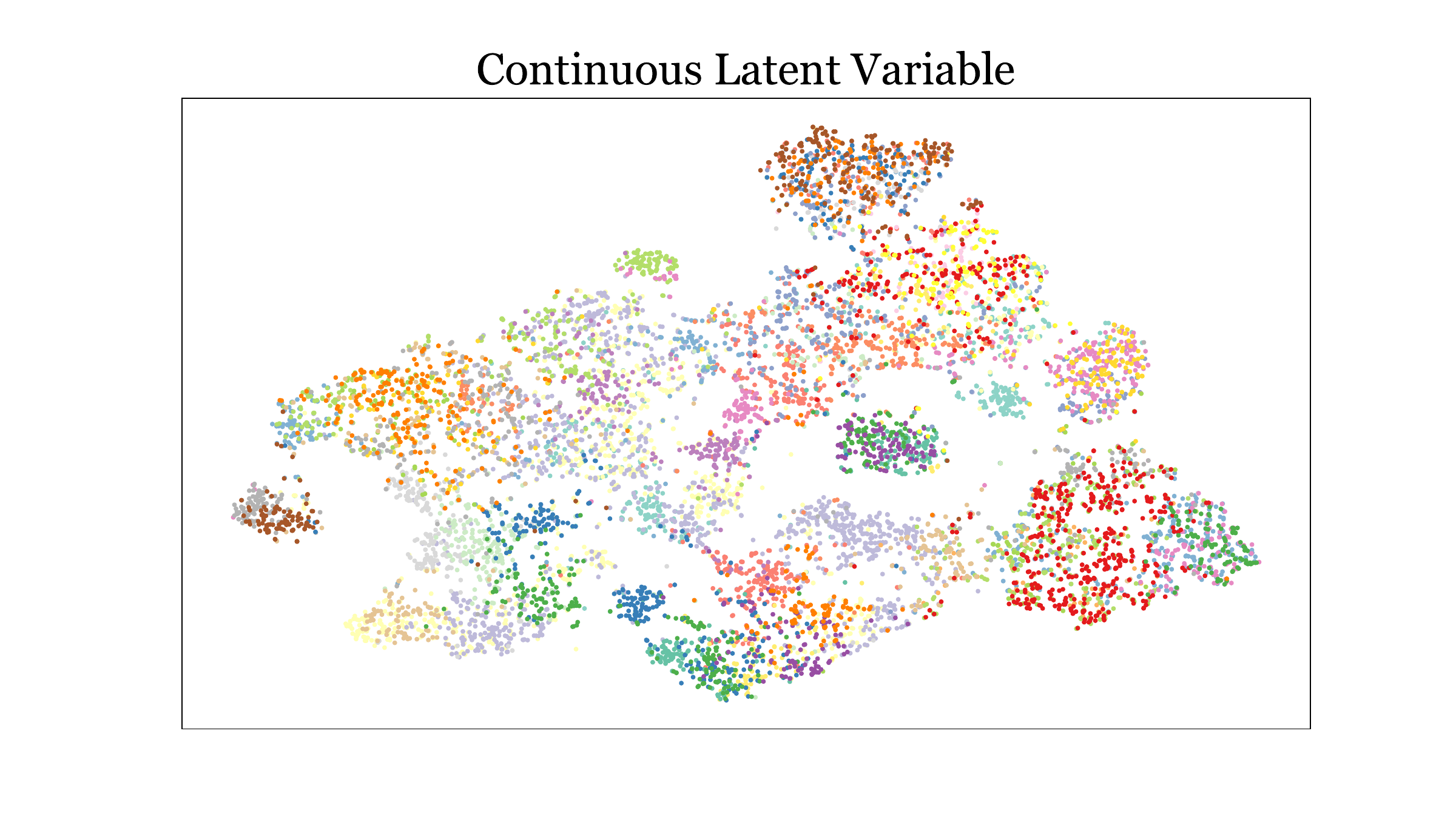}
    \caption{t-SNE visualization comparing question embeddings from GPT-2 with latent variable embeddings by VOLTA. Points of the same color depict embeddings from the identical context. VOLTA showcases diverse embeddings for each context, contrasting the deterministic nature of a vanilla LM.}
    \label{fig:visualization}
\end{figure*}

\subsection{Ablation Study}
\label{sec:ablation}

To assess the impact of the cross-attention-based latent-space connection, we compare VOLTA with Optimus in language modeling and dialog response generation. Given that QAG uniquely involves both continuous and discrete latent variables/codes, we focus on ablating the latent space information specifically within this task. Hence, the impact of VOLTA's three primary components is as follows:

\begin{itemize}
    \item \textbf{Cross-attention-based latent-space connection} (Table~\ref{tbl:lm},~\ref{tbl:dialog}): Optimus employs two distinct and intricate channels—embedding concatenation and summation—which pose challenges in optimization. In contrast, VOLTA's unified cross-attention-based approach is more stable. In language modeling, Optimus's latent variables even collapsed without its second-stage pretraining.
    \item \textbf{Latent variables} (Table~\ref{tbl:ablation}): Rows \texttt{A-D} show a general detriment when there is either an excess or a shortage of latent variables. Row \texttt{E} illustrates that when the latent variables become deterministic, the model essentially transforms into a conventional Autoencoder. Consequently, the performance experiences a significant decline, underscoring the critical role of the VAE framework.
    \item \textbf{Latent codes} (Table~\ref{tbl:ablation}): Rows \texttt{F, G} depict a further decline in performance when we further fix the latent codes from Row \texttt{E}, underlining the latent codes' role in enhancing generative diversity.
\end{itemize}

\subsection{Qualitative Analysis}
Table~\ref{tbl:latent-variable} exemplifies a diverse instance of QAG achieved by the variational nature of VOLTA. Our model architecture facilitates two more methods to alter the generation process. One method involves interpolating latent variables, detailed in Table~\ref{tbl:latent_var_interpolation}. The other method is centered on adjusting the InfoGAN-style latent codes, demonstrated in Table~\ref{tbl:latent-code}. In contrast to latent variables, latent codes are decoupled from the input context, affording the model more flexibility to explore the latent space.

To visualize the distribution of latent variables within the latent space, we utilize t-SNE \cite{van2008visualizing} to represent latent variable embeddings in a 2D space, comparing them with GPT-2 embeddings. Figure~\ref{fig:visualization} illustrates that GPT-2 produces identical embeddings for a given context. Conversely, our model displays the ability to generate a cluster of diverse Gaussian latent variable points of the same color, subsequently decoded into a spectrum of distinct questions.

\section{Conclusion}
We present VOLTA, a framework merging the power of Transformers with the variability inherent in VAE and InfoGAN. Diverging from prior approaches, VOLTA introduces a novel cross-attention-based connection linking the latent space to the decoder, enhancing stability in optimization. This innovative architecture accommodates diverse Transformer types, including decoder-only and encoder-decoder architectures, and supports varying latent variable types, whether continuous or discrete. Additionally, our framework incorporates InfoGAN-style latent codes, enabling input-independent variability, thereby further enriching generative diversity. Comprehensive experiments across six datasets spanning three distinct NLG tasks showcase VOLTA's significant enhancement in generative diversity while preserving quality.

\section{Limitations}

Given limited computational resources, we did not integrate LLMs into the VOLTA framework, leaving this as a potential area for future exploration. As our model architecture is not confined to GPT-2 or T5, larger and more robust Transformer models could be employed to demonstrate its generalizability. Additionally, incorporating more NLG tasks and datasets could further reinforce our experimental results.

\section*{Acknowledgements}
The research presented in this paper was partially supported by the Research Grants Council of the Hong Kong Special Administrative Region, China (CUHK 14222922, RGC GRF 2151185).

\bibliography{references}

\appendix

\clearpage
\section{Appendix}

\subsection{Notations}

Because we use VAE in this paper, our model is the composition of its encoder and decoder: $f = f_{enc} \circ f_{dec}$.

\begin{table}[h]
     \centering
     \resizebox{1\columnwidth}{!}{%
        \begin{tabular}{ c | p{2.5cm} p{6cm} }
         \toprule
         & Symbol & Description \\
         \hline
         \multirow{3}{*}{\rotatebox[origin=c]{90}{Constant}}
         & $m \big/ n$
            & The length of the context$\big/$question \\
         & \multirow{2}{*}{$k$}
            & The number of categories in a categorical distribution \\

        \hline
         \multirow{6}{*}{\rotatebox[origin=c]{90}{Variable}}
         & $\bm{x}$
            & Text sequence \\
         & $\bm{z} \big/ \bm{c}$
            & Latent variable$\big/$code vector \\
         & $z \big/ c$
            & A single latent variable$\big/$code \\
         & $\openbox_c \big/ \openbox_g \big/ \openbox_a$
            & Context$\big/$generation$\big/$answer subscript \\
         & $\openbox_{\square,i}$
            & Element index of a vector \\
         & $s \big/ e$
            & Answer span start$\big/$end token index \\

        \hline
         \multirow{6}{*}{\rotatebox[origin=c]{90}{Model}}
         & $f_{enc}(\cdot)$ $\big/$ $f_{dec}(\cdot)$
            & Encoder$\big/$decoder \\
         & $\text{FC}(\cdot)$
            & Single fully-connected layer \\
         & $\mathcal{N}(\cdot)$
            & Gaussian distribution \\
         & $\text{Cat}(\cdot)$
            & Categorical distribution \\
         & $\text{Uni}(\cdot)$
            & Uniform distribution \\
         & $[\mycdots, \mycdots]$
            & Concatenation operation \\

         \bottomrule
        \end{tabular}
     }
     \caption{Notations used in this paper.}
     \label{tbl:notations}
\end{table}

\subsection{Basic Definitions}

Information is defined as:
\begin{align*}
I(X) = -\log P(X) = \log \frac{1}{P(X)}.
\end{align*}

Entropy is defined as:
\begin{align*}
H(X) =& \mathbb{E}[I(X)] \\
=& \mathbb{E}[-\log f(X)]\\
=& - \int f(x)\log {f(x)} \mathrm{d}x,\\
H(X|Y) =& \mathbb{E}_{X, Y}[-\log f(X|Y)]\\
=& -\int\int f(x,y)\log f(x|y)\mathrm{d}x\mathrm{d}y,
\end{align*}
where $f(x)$ is the probability density function of a continuous distribution, and $f(x,y)$ is the joint probability density function.

Then the mutual information is:
\begin{align*}
& I(X;Y) \\
=& D_\text{KL}(P(X,Y) \parallel P(X)P(Y))\\
=& \int\int f(x,y)\log {\frac {f(x,y)}{f(x)f(y)}} \mathrm{d}x \mathrm{d}y \\
=&  - \int\int f(x,y)\log {f(y)} \mathrm{d}x \mathrm{d}y \\
 & + \int\int f(x,y)\log {\frac {f(x,y)}{f(x)}} \mathrm{d}x \mathrm{d}y\\
=& - \int f(y)\log {f(y)} \mathrm{d}y \\
 & + \int\int f(x,y)\log f(y|x) \mathrm{d}x \mathrm{d}y \\
=& H(Y) - H(Y|X)\\
=& H(X) - H(X|Y),
\end{align*}
because Kullback–Leibler divergence is defined to be:
\begin{align*}
& D_\text{KL}(Q \parallel P) \\
=& H(Q, P) - H(Q) \\
=& \mathbb{E}_Q[-\log p(X)] - \mathbb{E}_Q[-\log q(X)]\\
=& - \int q(x)\log p(x) \mathrm{d}x + \int q(x)\log q(x) \mathrm{d}x \\
=& \int q(x)\log {\frac {q(x)}{p(x)}} \mathrm{d}x \\
\geq & 0,
\end{align*}
where $H(Q, P)$ is the cross entropy of distributions $Q$ and $P$, with respective probability density functions $q(x)$ and $p(x)$.

\subsection{Optimus (Beta-VAE)}
\label{appendix:optimus}

In Optimus \cite{DBLP:conf/emnlp/LiGLPLZG20, DBLP:journals/corr/KingmaW13}, we assume a normal distribution for a continuous latent variable:
\begin{align*}
f(x) =& \frac{1}{ \sqrt{2\pi \sigma^2}   } e^{-\frac{1}{2}\left(\frac{x-\mu}{\sigma}\right)^2},\\
\log f(x) &\\
=& -\log (\sigma \sqrt{2\pi})  -\frac{1}{2}\left(\frac{x-\mu}{\sigma}\right)^2\\
=& -\log\sigma - \frac{1}{2}\log 2\pi -\frac{1}{2}\left(\frac{x-\mu}{\sigma}\right)^2\\
=& -\frac{1}{2}\log\sigma^2 - \frac{1}{2}\log 2\pi -\frac{1}{2}\left(\frac{x-\mu}{\sigma}\right)^2.
\end{align*}

We want $q(z|x)=\mathcal{N}(\mu_q, \sigma_q^2)$ and the prior, $p(z)=\mathcal{N}(\mu_p, \sigma_p^2) = \mathcal{N}(0,1)$, to be close
\begin{align*}
& D_\text{KL}(Q \parallel P)\\
=& -\int q(z)\log p(z)dz + \int q(z)\log q(z)dz \\
=& \left(\frac{1}{2}(\log 2 \pi \sigma_p^2) +\frac{\sigma_q^2 + (\mu_q - \mu_p)^2}{2\sigma_p^2}\right)  \\ & - \frac{1}{2}(1+\log 2 \pi \sigma_q^2)\\
=& \frac{1}{2}(\log \frac{\sigma_p^2}{\sigma_q^2}) +\frac{\sigma_q^2 + (\mu_q - \mu_p)^2}{2\sigma_p^2} - \frac{1}{2}\\
=& \frac{1}{2}\log \left(\frac{\sigma_p}{\sigma_q}\right)^2 +\frac{\sigma_q^2 + (\mu_q - \mu_p)^2}{2\sigma_p^2} - \frac{1}{2}.
\end{align*}

The mutual information between $z$ and $z|x$ is
\begin{align*}
I(z; x) =& H(z) - H(z|x),
\end{align*}
where the negative entropy for the normal distribution is:
\begin{align*}
 -H(z|x)
=& \mathbb{E}_{q(z|x)}[\log q(z|x)] \\
=&  \int q(z|x)\log q(z|x) \mathrm{d}z\\
=&  -\frac{1}{2}(1+\log(2 \pi \sigma_q^2))\\
=&  -\frac{1}{2}(1 + \log(2\pi)  + \log \sigma_q^2)\\
=& -\frac{1}{2}\log(2\pi) - \frac{1}{2}(1+\log \sigma_q^2),
\end{align*}
and the entropy of the normal distribution $q(z)$ is:
\begin{align*}
& H(z) =  \mathbb{E}_{q(z)} [-\log q(z) ]\\
=& - \int q(z) \left( -\log(\sigma_q \sqrt{2\pi}) -\frac{1}{2}\left(\frac{z-\mu_q}{\sigma_q}\right)^2 \right) \mathrm{d}z\\
=&   \int q(z)\log(\sigma_q \sqrt{2\pi}) \mathrm{d}z  \\
 & + \int q(z)\frac{1}{2}\left(\frac{z-\mu_q}{\sigma_q}\right)^2 \mathrm{d}z \\
=&  \mathbb{E}_{q(z)}[\log(\sigma_q \sqrt{2\pi})]  + \mathbb{E}_{q(z)}\left[\frac{1}{2}\left(\frac{z-\mu_q}{\sigma_q} \right)^2 \right]\\
=&  \log(\sigma_q \sqrt{2\pi}) + \frac{1}{2\sigma_q^2}\mathbb{E}_{q(z)}\left[(z-\mu_q)^2 \right]\\
=& \frac{1}{2}\log\sigma_q^2 + \frac{1}{2}\log (2\pi) + \frac{1}{2\sigma_q^2} (\sigma_q^2)\\
=& \frac{1}{2}\log\sigma_q^2 + \frac{1}{2}\log (2\pi) + \frac{1}{2},
\end{align*}
where $\mathbb{E}_{q(z)}\left[(z-\mu_q)^2\right]$ is simply the expected squared deviation from the mean, which by definition is the variance of the distribution, $\sigma_q^2$.

\subsection{Info-HCVAE}
\label{appendix:hcvae}
According to Info-HCVAE \cite{DBLP:conf/acl/LeeLJKH20}, some inputs are better suited to be encoded into discrete latent variables. In this case, we can make use of the categorical distribution:
\begin{align*}
P(x=i\mid {\bm{p}}) &= p_{i},
\end{align*}
where the event probabilities $\bm{p} = (p_{1},\ldots ,p_{k})$ and $\sum _{i=1}^{k}p_{i} = 1$; $k>0$ is the number of categories.

The Gumbel-Softmax distribution enables back-propagation through discrete distributions. The Gumbel distribution is:
\begin{align*}
\text{Gumbel}(\mu, \beta) = f(x; \mu, \beta) = \frac {1}{\beta }e^{-(z+e^{-z})},
\end{align*}
where $z={\frac {x-\mu }{\beta }}$.

To sample a category from the categorical distribution using the Gumbel-Max reparameterization trick, one can follow:
\begin{align*}
\argmax_{i} (G_{i} + \log p_{i}),
\end{align*}
where $G_{i} \sim \text{Gumbel}(0, 1)$. $\argmax$ can be made differentiable by approximating it with the softmax function:
\begin{align*}
y_i = \frac{\exp((G_{i} + \log p_{i})/\tau)}{\sum_{j=1}^{k} \exp((G_{j} + \log p_{j})/\tau)},
\end{align*}
where $\tau$ is the temperature parameter that controls the continuous relaxation.

Given two categorical distributions $P$ and $Q$, parameterized by $\bm{p}$ and $\bm{q}$, respectively, the KL divergence between them is:
\begin{align*}
D_\text{KL}(Q \parallel P) &= \sum_{i=1}^{k} q_{i} \log \frac{q_{i}}{p_{i}}.
\end{align*}

\subsection{InfoGAN}
\label{appendix:infogan}
The input noise $z$ is passed into the generator along with the latent code $c$: $G(z,c)$, where $z$ is concatenated with $c$. Because the generator can simply ignore the latent code $c$, InfoGAN \cite{DBLP:conf/nips/ChenCDHSSA16} adds Variational Mutual Information Maximization (VMIM) to maintain the mutual information between generated sample $x\sim G(z,c)$ and latent code $c$:
\begin{align*}
& I(c; G(z,c)) \\
=& H(c) - H(c|G(z,c))\\
=& H(c) + \mathbb{E}_{x\sim G(z,c)}[\mathbb{E}_{c'\sim P(c|x)}[\log P(c'|x)]]\\
=& H(c) + \mathbb{E}_{x\sim G(z,c)} \Big[ \sum_{c'}P(c'|x)\log P(c'|x) \Big]\\
=& H(c) + \mathbb{E}_{x\sim G(z,c)} \Big[ \sum_{c'}P(c'|x) \left(\log \frac{P(c'|x)}{Q(c'|x)} \right. \\
        & \hspace{3cm} \left. + \log Q(c'|x)\right) \Big]\\
=& H(c) + \mathbb{E}_{x\sim G(z,c)} \Big[ \sum_{c'}P(c'|x)\log \frac{P(c'|x)}{Q(c'|x)} \\
        & \hspace{3cm} + \sum_{c'}P(c'|x) \log Q(c'|x) \Big]\\
=& H(c) + \mathbb{E}_{x\sim G(z,c)} \big[ D_\text{KL}(P(\cdot|x) \parallel Q(\cdot|x)) \\
        & \hspace{3cm} + \mathbb{E}_{c'\sim P(c|x)}[\log Q(c'|x)] \big]\\
\geq& H(c) + \mathbb{E}_{x\sim G(z,c)} \left[ \mathbb{E}_{c'\sim P(c|x)}[\log Q(c'|x)] \right]\\
=& H(c) + \mathbb{E}_{c \sim P(c), x\sim G(z,c)} [\log Q(c|x)].
\end{align*}

Because the true posterior $P(c|x)$ is hard to obtain, an auxiliary distribution $Q(c|x)$ is added to approximate it, where $Q$ is parameterized by a neural network. Note that the final step applies the Law of Total Expectation (Lemma 5.1 in the InfoGAN paper), avoiding the need to sample from $P(c|x)$ and allowing the expectation to be computed directly by sampling from the prior $P(c)$. In practice, the entropy of the latent codes $H(c)$ is treated as a constant and omitted from the InfoGAN optimization objective.

\subsection{InfoVAE and InfoMax-VAE}
The evidence lower bound (ELBO) of a regular VAE is:
\begin{align*}
&\text{ELBO} \\
=& - \mathcal{L}_{\text{AE}}(\bm{x}) - \mathcal{L}_{\text{REG}}(\bm{z})\\
=& \mathbb{E}_{q_{\phi}(\bm{z}|\bm{x})}[\log p_{\theta}(\bm{x}|\bm{z})] - D_\text{KL}(q_{\phi}(\bm{z}|\bm{x}) \parallel p(\bm{z}))\\
\leq & \log p_{\theta}(\bm{x}).
\end{align*}

InfoVAE \cite{DBLP:conf/aaai/ZhaoSE19} and InfoMax-VAE \cite{DBLP:conf/isit/Lotfi-RezaabadV20} add mutual information to the loss:
\begin{align*}
\mathcal{L}(x) &= \mathcal{L}_{\text{AE}}(x) + \beta \mathcal{L}_{\text{REG}}(x) - \alpha I_q(x;z)\\
&= - \mathbb{E}_{p_{D}(x)}[\mathbb{E}_{q_{\phi}(z|x)}[\log p_{\theta}(x|z)]] \\
 &\quad + \beta \mathbb{E}_{p_{D}(x)} [D_\text{KL}(q_{\phi}(z|x) \parallel p(z))] \\
 &\quad - \alpha D_\text{KL}(q_{\phi}(x,z) \parallel q(x)q_{\phi}(z)).
\end{align*}

Because $D_\text{KL}(q_{\phi}(x,z)\parallel q(x)q_{\phi}(z))$ is usually intractable, it can be approximated with any one of the following:\\
\begin{itemize}
    \item KL divergence
    \item $f$-divergence (InfoMax)
    \item Donsker-Varadhan dual representation (InfoMax)
    \item Jensen-Shannon divergence (AAE)
    \item Stein Variational Gradient
    \item Maximum-Mean Discrepancy
\end{itemize}

\subsection{QA mutual information loss}
We want to enforce the mutual information (QAMI) between the generated QA pair. Following Info-HCVAE \cite{DBLP:conf/acl/LeeLJKH20}, we base this QAMI objective on Jensen-Shannon Divergence, which uses a bilinear layer on top of the decoder to classify whether the question and answer are a true pair:
\begin{align} \label{eq:QAMI}
\bm{h}_q ={}& \overline{\bm{h}}_{q,m+1:m+n} \quad \bm{h}_a = \overline{\bm{h}}_{a,1:m} \notag \\
g(q, a) ={}& \sigma({\bm{h}_q}^\mathsf{T} \bm{W} \bm{h}_a) \notag\\
I(q, a)  \geq{}&
\mathbb{E}[\log g(q, a)]
    \begin{aligned}[t]
        & + 1/2 \mathbb{E}[\log(1 - g(\tilde{q}, a))]\\
        & + 1/2 \mathbb{E}[\log(1 - g(q, \tilde{a}))]\\
    \end{aligned}   \notag \\
={}& - \mathcal{L}_{\text{QAMI}}(\bm{x}) ,
\end{align}
where the question and answer embeddings, $\bm{h}_q$ and $\bm{h}_a$, are the average of their contextualized token embeddings; $\bm{W}$ is the parameter matrix of the bilinear layer $g(\cdot)$; $\tilde{q} / \tilde{a}$ is a negative question$/$answer sample; $\sigma(\cdot)$ is the activation function.

\subsection{Examples}
We provide more examples of latent code control.

\begin{table}[h]
 \resizebox{1\columnwidth}{!}{%
    \begin{tabular}{ p{0.3cm} p{1.5cm} p{9.7cm} }
     \toprule
        \multirow{3}{*}{\rotatebox[origin=c]{90}{\textbf{}}} &
        \multicolumn{2}{p{11.2cm}}{
        \setlength\fboxsep{1pt}
        Knute Rockne became head coach in 1918. Under Rockne, the Irish would post a record of \colorbox{blue!30}{105} wins, 12 losses, and five ties \mycdots
        }
        \\
        \hline
        \textbf{Q1} & $c_g=-1$ &
            How many wins did Knute Rockne post?\\
        \textbf{Q2} & $c_g=-.9$ &
            How many wins did Knute Rockne have?\\
        \textbf{Q3} & $c_g=-.5$ &
            How many wins did the Irish post in 1918?\\
        \textbf{Q4} & $c_g=+.9$ &
            How many wins did the Irish post a record of in 1918?\\
     \bottomrule
    \end{tabular}
 }

 \resizebox{1\columnwidth}{!}{%
    \begin{tabular}{ p{0.3cm} p{1.5cm} p{9.7cm} }
     \toprule
        \multirow{3}{*}{\rotatebox[origin=c]{90}{\textbf{}}} &
        \multicolumn{2}{p{11.2cm}}{
        \setlength\fboxsep{1pt}
        The Lobund Institute grew out of pioneering research in germ-free-life which began in \colorbox{blue!30}{1928} \mycdots
        }
        \\
        \hline
        \textbf{Q1} & $c_g=-1$ &
            When did the institute begin research on germ free-life?\\
        \textbf{Q2} & $c_g=-.8$ &
            When did research in animal and plant life begin?\\
        \textbf{Q3} & $c_g=-.5$ &
            When did Lobund begin research on germ?\\
        \textbf{Q4} & $c_g=-.1$ &
            When did the Lobund Institute begin its research?\\
        \textbf{Q5} & $c_g=+.5$ &
            When did research in germ free-life begin?\\
     \bottomrule
    \end{tabular}
 }
    \caption{Examples of latent codes. Answers are highlighted in \colorbox{blue!30}{blue}. The latent code appears to control the specificity of the question.}
    \label{tab:more_lc_examples}
\vspace{-15pt}
\end{table}

\end{document}